# An Empirical Study on Position of the Batch Normalization Layer in Convolutional Neural Networks

Moein Hasani
Department of Computer Engineering, Bu Ali Sina
University, Hamedan, Iran
M.Hasani@eng.basu.ac.ir

Hassan Khotanlou
Department of Computer Engineering, Bu Ali Sina
University, Hamedan, Iran
Khotanlou@basu.ac.ir

*Abstract*—In this paper, we have studied how training of the convolutional neural networks (CNNs) can be affected by changing the position of the batch normalization (BN) layer. Three different convolutional neural networks have been chosen for our experiments. These networks are AlexNet, VGG-16, and ResNet-20. We show that the speed-up provided by the BN algorithm can be further improved by using the BN in positions other than the one suggested by its original paper. Also, we discuss how the BN layer in a certain position can aid the training of one network but not the other. Three different positions for the BN layer have been studied in this research, these positions are: BN layer between the convolution layer and the non-linear activation function, BN layer after the non-linear activation function and finally, the BN layer before each of the convolutional layers.

## I. Introduction

Normalizing the input of the neural networks has been proved to be advantageous to neural networks by increasing their learning speed. Batch normalization (BN) [1] extends this idea and normalizes the activations of intermediate layers in the network. This is attained by adding additional layers of the BN within a deep neural network. Normalization is performed across mini-batches and not the entire training set. BN effectiveness has been proved by many experiments during the last couple of years. Although, there is some disagreement in the machine learning community on what would be the appropriate position for the BN layer in a network to achieve the highest acceleration in the training process. In the original BN paper, authors suggest that the BN layer should be positioned before the non-linear activation function, although in practice there are some cases that show this position of the BN layer does not always result in the maximum speed-up in training process.

In this paper we aim to study how altering the position of the BN layer can affect the training duration, and how the position suggested for the BN layer in the original paper might not always be the most effective one. We have chosen three well-known convolutional neural networks (CNNs) for conducting our experiments. The selected networks are AlexNet [12], VGG [2] and ResNet [3]. We show that the speed-up provided by the BN algorithm can vary depending on its position in the network. Also, we discuss how an arrangement of layers can be useful for one network and not for the other. We test three different positions for the BN layer in our study. These positions are: BN layer between the convolution (Conv) layer and the non-linear activation function, BN layer after the activation function and BN layer right before each of the Conv layers.

Batch normalization algorithm

The BN algorithm for CNNs works as follows: for each channel, the mean and standard deviation of the activations in mini-batch are calculated, then the calculated standard deviation is subtracted from the activations. The result of subtraction is divided by the square root of variance plus some ε value that is used to prevent the division by zero. The BN uses two trainable parameters $\gamma$ and $\beta$, so the effect of normalization can be controlled by the optimizer. If $\gamma$ equals the square root of variance plus ε and $\beta$ equals the mean of the activations in the mini-batch the normalization can be undone. During the test time, averages of the mean and the standard deviation which were obtained during the training are used. Equation (1) shows how the activations in a mini-batch are normalized by the BN. $I_k$ is the activations in dimension k of the mini-batch, $\mu_k$ is the mean of activations in dimension k of the mini-batch, and $\sigma_k^2$ is the variance of activations in dimension k of the mini-batch.

$$O_k = \gamma_k \frac{I_k - \mu_k}{\sqrt{\sigma_k^2 + \varepsilon}} + \beta_k \qquad (1)$$

## II. Networks and Arrangements

### A. Arrangements

In this paper, we study three different ways of arranging the layers in CNNs with the BN layer. The first way of arranging the layers is similar to the one that the BN paper suggested originally. The BN layer is placed after the Conv layer and normalizes the activations before passing them to the non-linear activation function. We test two more ways of arranging the layers within a network. In the first one, the BN layer is positioned after the non-linear activation function and in the other, the BN layer is before the Conv layer and it normalizes the input (except for the first Conv layer of the network). These arrangements have been depicted in Fig. 1.

Furthermore, we use the BN layer between the fully connected (FC) layers of our models. In arrangement 1, the BN layer is positioned after each FC layer and before the non-linear activation function, and in arrangements 2 and 3 the BN layer is placed after the non-linearity. Moreover, we place the BN layer before the dropout [10] in our work.

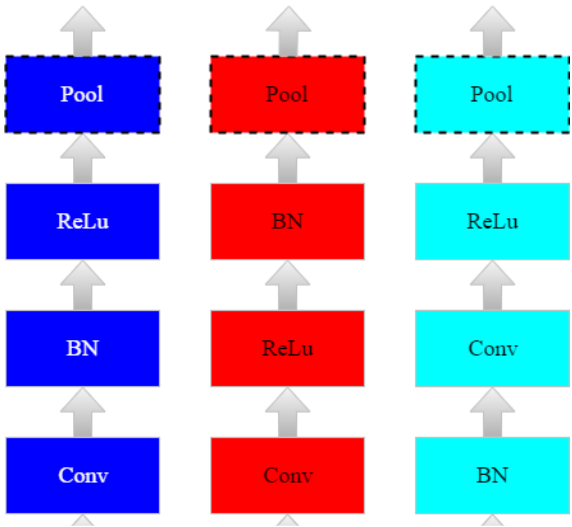

Fig. 1. Different arrangements of layers used in this study. (a) arrangement 1, (b) arrangement 2 and (c) arrangement 3. (The Pooling layer is dashed because it might not always be present after the third layer in these arrangements.)

### B. Networks

#### 1) AlexNet

AlexNet is known as the first deep CNN architecture. It was proposed by Krizhevesky et al. [13] and managed to achieve the state of the art results in ImageNet Large Scale Visual Recognition Competition 2012 (ILSVRC 2012). This network has more layers and parameters compared to CNNs prior to it like LeNet-5 [4]. The design of many other CNNs after AlexNet has been inspired by the depth of this architecture and its efficient learning approach. This architecture uses ReLU [8] as a non-saturating activation function to cope with the problem of vanishing gradient [6]. Overlapping sub-sampling, local response normalization (LRN) and dropout [12] were used in order to prevent the over-fitting problem in this architecture [12]. AlexNet uses large size filters (11x11 and 5x5) at its initial layers which had not been practiced in CNNs before it. In the early days of introducing AlexNet, it was trained on 2 GPUs to overcome the hardware shortcomings. However, in this research, we trained this network on only one GPU. Additionally, the LRN layers in the original AlexNet architecture are replaced by the BN layers in this study. This network has been tested with the three arrangements introduced earlier in this paper.

#### 2) VGG-16

Inspired by the extraordinary results achieved by the CNNs, Simonyan and Zisserman proposed a simple architecture for designing CNNs. This new architecture was named VGG and it is famous for its simple and homogenous architecture. VGG came as the 1st runner-up in ILSVRC 2014 and showed the state of the art result in the localization task. This architecture demonstrated that replacement of 11x11 and 5x5 filters with 3x3 ones can have the same effect of the large size filters and provides a low computational complexity by reducing the number of parameters. These findings encouraged the researchers to work with smaller size filters. For decreasing the computational cost, VGG uses max-pooling [7] after Conv layers and padding to preserve the input size. Also, it takes advantage of 1x1 convolution in order to decrease the complexity of the network. VGG has shown excellent results in image classification tasks and localization problems. Although, it has one major drawback and it is its high computational cost, which is due to the large number of layers used in this network. Even though the BN algorithm has increased the learning speed of VGG, it is still relatively slower than other networks like AlexNet. In this study, we use the VGG-16 architecture from the original paper and like the AlexNet, the training process of this network is evaluated with all three of the introduced arrangements.

#### 3) ResNet-20

ResNet was proposed by He et al [3]. The authors used 152-layers deep residual network in the ILSVRC 2015 and managed to win the competition. ResNet is many times deeper than previously proposed architectures and it shows less computational complexity. ResNet employs a technique called skip connection that causes fewer layers to be propagated through in backpropagation in order to speed-up the learning process and reduce the effect of vanishing gradients [6]. ResNet also attained a 28% improvement image recognition benchmark dataset COCO [9]. Groundbreaking performance of ResNet on computer vision tasks illustrated the important role that the depth of a network plays in its success. We selected ResNet-20, which is one of the proposed architectures in the original ResNet paper, for our experiments in this research. There is only one pooling layer in this architecture and it is placed before the first FC layer.

## III. RESULTS

### A. Experimental setup

To investigate the results of utilizing the BN layer in different positions, we use three image datasets: CIFAR10, CIFAR100 and Tiny ImagNet [11].

Each of the CIFAR10 and CIFAR100 has 60000 32x32 color images. CIFAR10 dataset consists of 10 classes of 6000 images and the CIFAR100 dataset includes 100 classes of each 600 images. We use 50000 of images from each of the CIFAR10 and CIFAR100 for the training process.

Tiny ImageNet is a subset of ImageNet dataset that contains 100000 64x64 color images which are composed of 200 classes each with 500 samples. We resize the images in this dataset to 32x32. All the 100000 images of Tiny ImagNet have been used for the training.

A batch size number of 512 is used in the training of the networks. The state of the art optimizer, Adam [5], has been adopted as our chosen optimizer with the initial learning rate of 0.001 and β1 of 0.9 and β2 of 0.999. Tensorflow [12] library has been used for implementing the networks and algorithms of this paper.

## B. Training results

Here we compare the effects that the three different arrangements introduced earlier have on the training process of the chosen networks in terms of the steps (number of the batches of the data seen by the network) each arrangement takes. AlexNet and VGG-16 networks are tested with all the three different arrangements but, ResNet-20 is tested only with the first two arrangements since there is only one pooling layer in ResNet-20. Due to the limitation of computational resources, certain values are chosen as the acceptable accuracy for each dataset. Tables show how many training steps have been taken by each arrangement of the networks to reach the selected accuracy values for each dataset. The full training process of the networks is depicted in figure (2).

### 1) Training results of AlexNet

According to the results of training the AlexNet on the selected datasets (Table I), by positioning the BN layer before the Conv layers (arrangement 3), AlexNet can be trained faster than other ordering of the layers. To reach the specified accuracy values, AlexNet with arrangement 3 has taken almost 43% fewer training steps on CIFAR10, 53% fewer training steps on CIFAR100 and 33% fewer training steps on Tiny ImageNet compared to arrangement 1 that is proposed by the original paper. On the contrary to the speed-up provided by arrangement 3, positioning the BN layer right after the non-linear activation function (arrangement 2) has caused the training of AlexNet to be longer compared to results of the training the AlexNet with arrangement 1.

### 2) Training results of VGG-16

Results from training the VGG-16 on the datasets (Table II) display that the appropriate position for the BN layer when training the VGG-16 depends highly on the dataset that this network is trained on. When we train the VGG-16 on a small and relatively easy to learn dataset like CIFAR10, the number of steps taken for reaching a decent accuracy is almost the same for all three of the arrangements. On the other hand, reports from training the network on CIFAR100 and Tiny ImageNet display a disagreement on the arrangement that provides the highest learning speed. Although arrangement 3 achieves the most speed-up when learning from CIFAR100 with an almost 50% faster training compared to the other arrangements, it doesn't repeat its success when we train the network on the more complex dataset, Tiny ImageNet. Additionally, arrangement 1 which was inferior to arrangement 3 when training the VGG-16 on CIFAR100 dataset, performs better than other arrangements on the Tiny ImageNet dataset. Furthermore, we observe that arrangement 2 has a lower performance compared to other arrangements when training on CIFAR100 and Tiny ImageNet.

### 3) Training results of ResNet-20

Studying the results from training ResNet-20 on CIFAR10 and CIFAR100 (Table II) shows that the two arrangements of this network attain very close results on these datasets. Tough, training this network on the Tiny ImageNet dataset clearly exhibits that arrangement 1 has done much better, and the network with this arrangement has taken almost 36% fewer steps to reach the same accuracy as the network with arrangement 2. In the case of ResNet-20, we can observe that the suggested position by the authors has been proved to be the most promising of two arrangements in terms of speeding up the learning process in our study.

TABLE I. NUMBER OF STEPS TAKEN BY DIFFERENT ARRANGEMENTS OF ALEXNET UNTIL THE SELECTED THRESHOLDS

| Arrangement | Datasets | | |
|---|---|---|---|
| | CIFAR10 95% accuracy | CIFAR100 90% accuracy | Tiny ImageNet 85% accuracy |
| #1 | 2.8 K | 4.8 K | 15.4 K |
| #2 | 6.1 K | 6.6 K | 21.6 K |
| #3 | **1.6 K** | **2.3 K** | **10.2 K** |

TABLE II. NUMBER OF STEPS TAKEN BY DIFFERENT ARRANGEMENTS OF VGG-16 UNTIL THE SELECTED THRESHOLDS

| Arrangement | Datasets | | |
|---|---|---|---|
| | CIFAR10 95% accuracy | CIFAR100 90% accuracy | Tiny ImageNet 85% accuracy |
| #1 | 1.4 K | 6.5 K | **13.6 K** |
| #2 | **1.1 K** | 6.5 K | 15 K |
| #3 | 1.2 K | **3.3 K** | 18.9 K |

TABLE III. NUMBER OF STEPS TAKEN BY DIFFERENT ARRANGEMENTS OF RESNET-20 UNTIL THE SELECTED THRESHOLDS

| Arrangement | Datasets | | |
|---|---|---|---|
| | CIFAR10 95% accuracy | CIFAR100 90% accuracy | Tiny ImageNet 85% accuracy |
| #1 | 2.4 K | **5.7 K** | **35.1 K** |
| #2 | 2.4 K | 6.2 K | 54.6 K |

## IV. CONCLUSION AND RECOMMENDATION

In this study, we have examined the result of adopting the BN layer in different positions in CNNs. The goal of our study has been to prove that employing the BN layer in other positions than the suggested one can lead to better or close results. For AlexNet, as results have displayed, using the BN layer before each Conv layer causes the model to learn faster than positioning it between the Conv layer and the non-linear activation function. Although, the results from the experiments on AlexNet have shown that arrangement 3 accelerates the learning ability more than two other ways of arranging the layers, this arrangement when applied to VGG-16 produces different results. Moreover, the results from experiments with ResNet-20 show that using the BN layer between the Conv layer and the non-linear activation function is preferable to the other arrangement for this network.

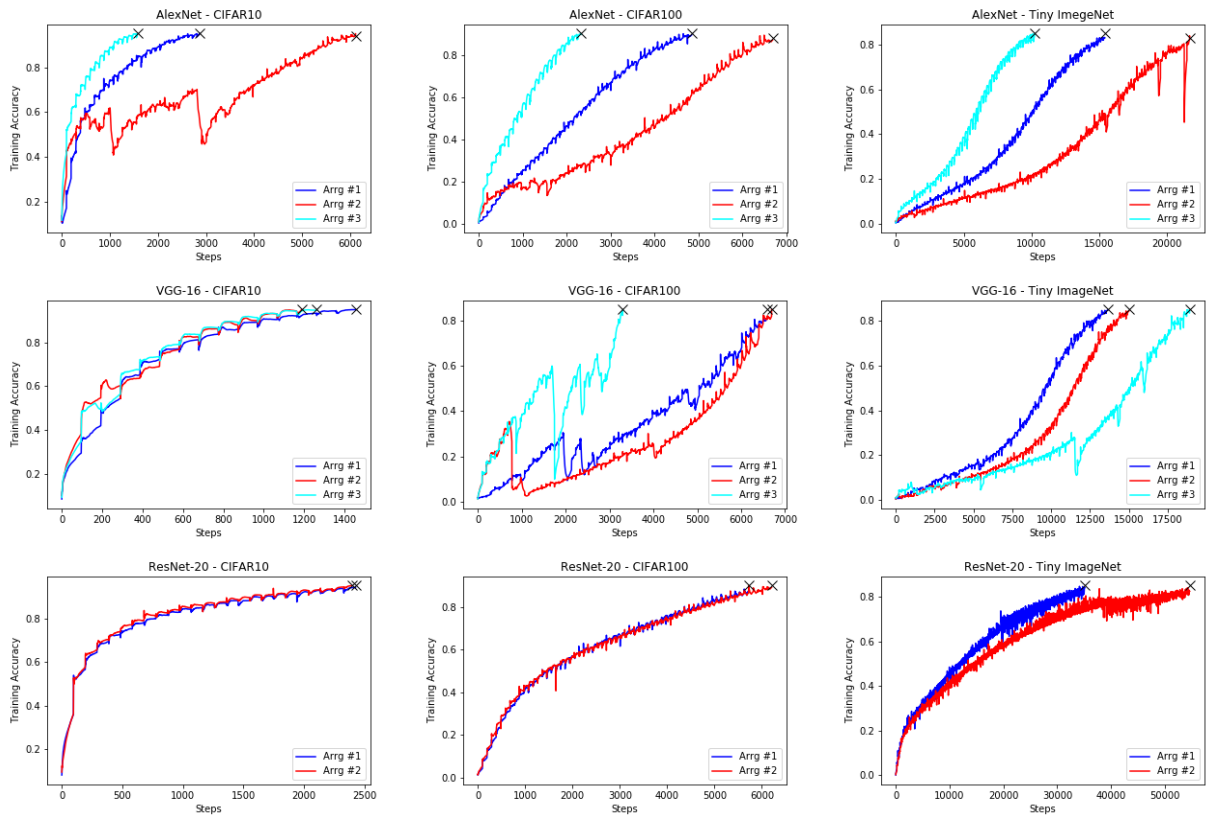

Fig. 2. *Illustration of change in accuracy with respect to the training step for different arrangements of the selected networks. Each figure shows the training process of a network with all of the arrangements.*

Even though the results provided in this research show the advantages of one arrangement over another, we cannot conclude that there is an absolute most effective way of ordering the layers that can be used in all CNNs. The results are variant when we apply these arrangements to different networks. We suggest training your selected network with all the three arrangements for a few epochs and check which arrangement causes the network to learn fastest. You can continue the training of the network with that arrangement.

We believe that these findings demonstrate that we should always search for a more beneficial way of adopting the techniques and algorithms in machine learning problems. The suggested ways of using the algorithms may not always turn out to be the most effective ones. The more efficient way of using the methods can only be revealed by doing various experiments.